\documentclass{article}
\usepackage{ijcai22}

\usepackage{times}

\usepackage{soul}
\usepackage{url}
\usepackage[hidelinks]{hyperref}
\usepackage[utf8]{inputenc}
\usepackage[small]{caption}
\usepackage{graphicx}
\usepackage{amsmath}
\usepackage{amsthm}
\usepackage{booktabs}
\usepackage{algorithm}
\usepackage{algorithmic}
\usepackage{natbib}
\usepackage{bbm,amssymb}
\usepackage{multirow}
\usepackage{verbatim}
\usepackage{array}
\usepackage[mathscr]{euscript}
\usepackage{color}
\usepackage{natbib}  % DO NOT CHANGE THIS AND DO NOT ADD ANY OPTIONS TO IT
\usepackage{bbding}
\usepackage[normalem]{ulem}

\newcommand{\tabincell}[2]{\begin{tabular}{@{}#1@{}}#2\end{tabular}}
\newcommand\blfootnote[1]{%
	\begingroup
	\renewcommand\thefootnote{}\footnote{#1}%
	\addtocounter{footnote}{-1}%
	\endgroup
}

\title{Automatically Generating Counterfactuals for Relation Classification}

\author{
Mi Zhang$^1$
\and
Tieyun Qian$^1$$^*$
\and 
Ting Zhang$^1$
\affiliations
$^1$School of Computer Science, Wuhan University, China\\
\{mizhanggd,qty,tingzhang\_17
\}@whu.edu.cn,
}

\begin{document}
	
	\maketitle
	
	\begin{abstract}
		The goal of relation classification (RC) is to extract the semantic relations between/among entities in the text. As a fundamental task in natural language processing, it is crucial to ensure the robustness of RC models. %Recently deep learning models have made substantial progress in RE tasks. Despite the high accuracy these models have achieved, they might be threatened by spurious correlations. One solution to this problem is to train the model with counterfactually augmented data (CAD) such that it can learn the causation rather than the confounding. While CAD have shown their benefits on improving the model robustness in  NLP area, the attempts have yet to be made for RE tasks.
		Despite the high accuracy current deep neural models have achieved in RC tasks, they are easily affected by spurious correlations. One solution to this problem is to train the model with counterfactually augmented data (CAD) such that it can learn the causation rather than the confounding. However, no attempt has been made on generating counterfactuals for RC tasks.
		
		In this paper, we  \emph{formulate the problem of automatically generating CAD for RC tasks} from an entity-centric viewpoint, and \emph{develop a novel  approach to derive contextual counterfactuals} for entities. Specifically, we exploit two elementary topological properties, i.e., the centrality and the shortest path, in syntactic and semantic dependency graphs, to first identify and then intervene on the contextual causal features for  entities. We conduct a comprehensive evaluation on four RC datasets by combining our proposed approach with a variety of backbone RC models. The results demonstrate that our approach not only improves the performance of the backbones, but also makes them more robust in the out-of-domain test.
	\end{abstract}
	\blfootnote{*Corresponding author.}
	\section{Introduction}
	Relation classification (RC) aims to extract the semantic relations between/among entities in the text. It serves as a fundamental task in natural language processing (NLP), which facilitates a range of downstream applications such as knowledge graph construction and question answering.
	Deep neural models have  made substantial progress  in many communities like  computer vision and NLP. However, existing studies~\citep{abs-1711-11561,Zhou_cvpr16} have shown that neural models are likely to be unstable due to the spurious correlations. A typical example is the dog on the grass or in the sea, where the head of the dog is the causation and the grass is a confounding.
	The robustness and generalization of the models can be severely affected by spurious correlations. Consequently, it is desirable to train robust models by identifying what features would need to change for the model to produce a specified output (label).
	
	One solution to train a robust neural model is to generate counterfactually augmented data (CAD)~\citep{KaushikHL20} such that the model can distinguish causal and spurious  patterns.
	There has been an increasing interest in generating counterfactuals in many subfields of NLP, including sentiment classification~\citep{Yang0CZSD20,ChenXY21,WangC21}, named entity recognition~\citep{ZengLZZ20}, dialogue generation~\citep{ZhuZLW20}, and cross-lingual understanding~\citep{YuZNS020}.
	Early research often employs human annotators and designs human-in-the-loop systems~\citep{KaushikHL20,SrivastavaHL20}. Most of recent studies automatically generate counterfactuals with  semantic interventions using templates, lexical and paraphrase changes, and text generation methods~\citep{MadaanPPS21,FernP21,RobeerBF21}, and a few of them incorporate the syntax into language models~\citep{YuZNS020}.
	
	Despite the emerging trend of casual analysis in the NLP field, automatically generating CAD for RC tasks has received little attention. The key challenge is that the RC task involves two or more entities which should remain unchanged during the intervention, otherwise the problem itself also changes. Existing  semantic intervention methods tend to select content words for generating counterfactuals. The reason is that they follow ``the minimal change'' principle~\citep{KaushikHL20,Yang0CZSD20}, where the content words like entities  have a larger probability to be replaced since they convey more information than function words. This is not desirable in our task.
	The syntactic intervention method~\citep{YuZNS020} replaces  each type of dependency relation between two words with a randomized one. Such a method cannot flip the label in our RC task, either.
	
	In view of this, we \emph{introduce the problem of automatically generating CAD into RC tasks for the first time}. To meet the condition of invariant entities, we formulate it from an entity-centric viewpoint. We then develop a novel approach to derive counterfactuals for the contexts of entities. Instead of directly manipulating the raw text,  we deploy semantic dependency graph (SemDG) and syntactic dependency graph (SynDG)  as they contain abundant information. We \emph{exploit two elementary topological properties to identify contextual casual features for entities}. In particular, the centrality measures the importance of the word in the SynDG, which helps us recognize structurally similar entities in two samples. Meanwhile, the shortest path between two entities in SemDG captures the basic relation between them. We then generate CAD by intervening on the contextual words around one specific entity and those along the shortest path between two entities.
	
	The contributions of this study are summarized as follows.
	
	$\bullet$ To the best of our knowledge, we are the first to investigate the problem of automatical generation of CAD in RC tasks, which can improve the model robustness  and is an essential property for real applications.
	
	$\bullet$ We propose a novel approach which exploits the topology structures in both the semantic and syntactic dependency graphs  to generate more human-like counterfactuals for each original sample.
	
	$\bullet$ Extensive experiments on four benchmark datasets prove that our approach significantly outperforms the state-of-the-art baselines. It is also more effective for alleviating spurious associations and improving the model robustness.

	\section{Related Work}
	\paragraph{Relation Classification} Deep learning models have been successfully employed in RC tasks, either for extracting better semantic features from word sequences~\citep{ZhouSTQLHX16,ZhangZCAM17}, or incorporating syntactic features over the dependency graph~\citep{GuoZL19,MandyaBC20}. More recently, pre-trained language model (PLM) based models have become the mainstream~\citep{QinLT00JHS020,YamadaASTM20}.
	
	Despite the remarkable performance deep neural models have achieved in RC, their realization in practical applications still faces big challenges. One particular concern is that these models might learn unexpected behaviors that are associated with spurious patterns.
	
	\paragraph{Counterfactual Reasoning}
	There has been a growing line of research to learn casual associations using casual inference. Early work attempts to achieve model robustness with the help of human-in-the-loop systems to generate counterfactual augmented data~\citep{KaushikHL20,SrivastavaHL20}. Recently, automatically generating counterfactuals has received more and more attention. For example, ~\cite{WangC21,Yang0CZSD20} identify causal features and generate counterfactuals by substituting them with other words for sentiment analysis and text classification. ~\cite{YuZNS020} generate counterfactually examples by randomly replacing syntactic features to implicitly force the networks to learn semantics and syntax. Another line of work deploys PLMs to obtain a universal counterfactual generator for texts~\citep{FernP21,RobeerBF21,MadaanPPS21,WuRHW20,TuckerQL21}.
	
	Overall, the problem of automatical generation of CAD has not been explored in RC tasks. Our work makes the first attempt on it. Furthermore, our proposed framework takes advantage of both syntax and semantic information in dependency graphs, which allows us to generate grammatically correct and semantically readable counterfactuals.

	\section{Preliminary}
	\subsection{Task Definition}
	\paragraph{Relation Classification (RC)}
	Relation classification tasks are mainly categorized into three types: sentence-level RC, cross-sentence $n$-ary RC, and document-level RC. Our proposed model is targeted for the first two types and can be extended to the last one which we leave for the future work.
	%~\footnote{There are several studies towards document-level RE where the relations are expressed across multiple paragraphs, and extra efforts are required to deal with the multiple mentions of an entity. Its learning focus is different from that of sentence-level and cross-sentence n-ary RE and thus we omit it.}..
	Formally,
	let $\mathcal{X}$ = [$t_{1}$, ..., $t_{m}$] be a text consisting of sentence(s) with $m$ tokens  and two or three entity mentions. $\mathcal{Y}$ = \{$y_{1}$, ..., $y_{p-1}$, $y_{p}$\}~($y_{p}$ = None) is a predefined relation set.
	The RC task can be formulated as a classification problem of determining whether a relation holds for entity mentions.
	\paragraph{Generating Counterfactuals in RC}
	Given a RC dataset \{($X$,$Y$)\} and the model \emph{f}: $X$$\to$$Y$, we aim to generate a set of counterfactuals \{($X_{\text{cf}}$,$Y_{\text{cf}}$)\} ($Y_{\text{cf}}$$\neq$$Y$) without changing entities. For the purpose of training a robust model, it is desirable to make $X_{\text{cf}}$ as similar to $X$ as possible~\citep{Yang0CZSD20}, i.e., $X_{\text{cf}}$ is syntactic-preserving and semantic-reasonable.
%(\textit{$\mathbb{X}$,$\mathbb{Y}$}) = \{{($X_1$,$Y_1$), \dots, ($X_n$,$Y_n$)}\}
%we aim to generate a set of counterfactuals \{($X_{\text{cf}}$, $Y_{\text{cf}}$)\} ($X_{\text{cf}}$ is as similar  to $X$ as possible and $Y_{\text{cf}}$ $\neq Y$ for training a robust model) without changing entities.
	%, which will be used as augmented data to improve the robustness of the model \emph{f}.
	%Given a text sample with its label, i.e., `Entity-Origin' relation text, a counterfactual sample is generated by our counterfactual generation method to the text in order to change its label, i.e., to `Product-Producer' relation.
	%Formally, given a RE dataset , and a trained RE model , our propose is to generate counterfactual sample $X_{\text{cf}}$ such that \textit{RE}($X_{\text{cf}}$) $\neq Y$.
	
	% More specifically, sentence-level RE focuses on the binary relations where two entities occur in the same sentence (i.e., $n$ = 2 and \textit{T} is a sentence) while cross-sentence n-ary RE focuses on multiple sentences with entities (i.e., $n > $  2 and \textit{T} contains multiple sentences).
	
	\subsection{Structural Causal Model}
	We introduce the structure casual model (SCM)~\citep{Judea2000} to investigate the casual relationship between data and the RC model, where random variables are vertices and an edge denotes the direct causation between two variables. Before start, we first propose our \emph{casual questions to guide the generation of SCM}.
	(1) What would happen if the important syntactic structure  \emph{SY} around the entity is changed?
	(2) What would happen if the semantic path \emph{SE} between two entities in the text changes?
	%These are casual questions that used to generate counterfactual data rather than original data alone.
	%~\cite{Pearl2009CausalII}.
	
\begin{figure}[htb]
		\vspace{-1mm}
		\center{\includegraphics[width=0.45\textwidth]{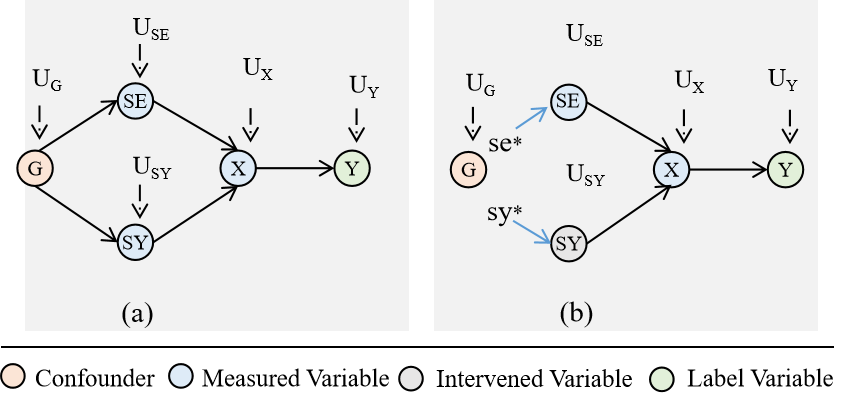}}
		\vspace{-2.5mm}
		\caption{An illustration of structural causal models (SCMs) that describes the mechanism of the causal inference for RC models.}
		
		\label{fig:scm}
		\vspace{-1.5mm}
	\end{figure}

	Based on the casual questions, we consider two important factors $SY$ and $SE$ as variables in SCM to capture the casual relation in the text. As shown in Figure~\ref{fig:scm}(a), there is a confounding variable $G$ that influences the generation of  $SY$ and $SE$, $Y$ is the label variable, and $U_{*}$ represents the unmeasured variable. $X$ $\to$ $Y$ means there exits a direct effect from $X$ to $Y$. Furthermore, the path $SY$/$SE$ $\to$ $X$ $\to$ $Y$ denotes $SY$/$SE$ has an indirect effect on $Y$ via a mediator $X$. $Y$ can be calculated from the values of its ancestor nodes, which is formulated as: $Y_{sy/se, x} = f(SY = sy/SE = se, X = x)$, where $f(.)$ is the value function of $Y$.
	
	%As shown in Figure~\ref{fig:scm}, a text \text{X} is composed by two important factors: syntax \text{SY} and semantics \text{SE}. We consider them as two variables in SCM to help us better understand the casual relationship in a text.
	
	\begin{comment}
	The casual effect of $SY$ on $Y$ is the magnitude by which the target variable $Y$ is changed by a unit intervention by fixing the value of \text{sy} as \text{sy}$^{*}$. This intervention blocks the influence of the variable $G$ on the variable $SY$.  Similarly, we also intervene on the variable $SE$ to estimate the casual effects of the variable $SE$, as shown in Figure~\ref{fig:scm}(b).
	
	The average casual effect of \text{SY} = \text{sy}$^*$ on \text{Y} is defined as:
	\begin{equation}
	\setlength{\abovedisplayskip}{6pt}
	\setlength{\belowdisplayskip}{6pt}
	\small
	\text{ACE} = \textbf{E}(\text{Y}_{\text{sy},\text{X}_{\text{sy}}}) - \textbf{E}(\text{Y}_{\text{sy}^{*},\text{X}_{{\text{sy}}^{*}}})
	\end{equation}
	which can be understood as the difference betweenn two hypothetical situations \text{SY}=\text{sy} and \text{SY}=\text{sy}$^*$. \text{SY}=\text{sy}$^*$ refers to a situation where the value of \text{SY} is muted from the reality, denoted as the intervention for RE models. \text{X}$_{{\text{sy}}^{*}}$ denotes the value of \text{X} when \text{SY}=\text{sy}$^*$.
	\end{comment}

	In order to  estimate the casual effects of the variable $SY$/$SE$ on  the target variable $Y$, we need to block the influence of the confounding variable $G$ on $SY$/$SE$, and see how $Y$ is changed by a unit intervention when fixing the value of $sy$/$se$ as $sy$$^{*}$/$se$$^{*}$, as shown in Figure~\ref{fig:scm}(b).

	\section{Model Overview}
	%Our approach is a two-stage process: we identify causal features with backbone encoder to generate counterfactuals and train a robust classifier using CAD.
	This section presents our approach for automatically generating counterfactuals by substituting casual compositions with candidate ones to enhance the robustness of RC models.
	One of our key insights is adopting the graph formulation to incorporate the rich information in syntactic dependency graph (SynDG) and semantic dependency graph (SemDG).
	%By adopting the graph formulation, our framework subsumes prior approaches based on SemDG and SynDG to incorporate a rich set of linguistic and semantics analyses to aid RE.

	SynDG pays  attention to the role of \emph{non-substantive words} such as prepositions in the sentence. Existing work~\citep{GuoZL19} has shown the effectiveness by applying SynDGs to RC models with graph convolutional networks. We also exploit SynDG in this study but our goal is to identify  causal syntactic features.
	Moreover, we introduce  SemDG into the field of causality analysis. Our intuition is that  SemDG reveals the semantic relationship between \emph{substantive words}, and crosses the constraints of the surface syntactic structure. In other words, SemDG provides complementary information for SynDG.
	
	%Early methods generally exploit syntactic features in SynDG such as dependency relation and POS tags to extract relation between entities. And SynDG pays more attention to the role of non-substantive words (such as prepositions) in syntactic structure. Meanwhile, SemDG analyzes the semantic relationship between each word unit of the text, and crosses the constraints of the sentence surface syntactic structure and directly obtain deep semantic information.	Semantic dependence tends to establish a direct dependence arc between substantive words with direct semantic correlation, and non-substantive words exist as auxiliary markers.
	
	A counterfactual, denoted as $X_{\text{cf}}$,  is a sample which has the most similar semantic or syntactic structure with the original sample $X_{\text{ori}}$ but has a different label. Recall that we cannot change the entities during the interventions for samples in RC tasks, hence we propose our \textbf{entity-centric framework} to generate counterfactuals  by first identifying and then intervening on contextual casual features for entities via topological based analysis in SynDG and SemDG. The identification of causal features consists of two main steps.
	\begin{itemize}
		\item To identify \textbf{the syntactic casual composition around the entity}, we conduct the centrality analysis for entities in two samples since centrality measures how importance a node is in the graph. %We then substitute the neighbours of the similar entities in these samples.
		
		\item To identify \textbf{the semantic casual composition between two entities}, we employ the shortest dependency path (SDP) between them since SDP retains the most relevant information while eliminating irrelevant words (noises) in the sentence~\citep{XuMLCPJ15}.
	\end{itemize}

	\subsection{Generating Syntactic Counterfactuals}
	Since a counterfactual  should flip the label of the original sample $X_{\text{ori}}$, the candidate substitute $X_{\text{can}}$ which will be used for identifying causal features is randomly chosen from the training samples with different class labels~\footnote{In our implementation, we randomly select three candidate substitutes for the original sample. If none of  generated candidates meets the requirements, they will all be discarded. If there are multiple candidates qualified as counterfactuals, one of them is chosen at random.}. Moreover, since entities are of the most importance in RC tasks, the entities in the substitute should be most similar with those in $X_{\text{ori}}$. In view of this, we first identify the syntactically similar entity nodes  in SynDG using the centrality metric, which is proposed to account for the importance of nodes in a graph (network). We employ three types of centralities including betweenness centrality (BC), closeness centrality (CC), and degree centrality (DC) for this purpose. After that, we generate the contextual counterfactual for these syntactically similar entities to meet the condition of invariant entities, i.e., instead of changing entities, we intervene on their contexts.
	We term this proposed method \textbf{SynCo} as the substitution of casual features is based on the syntactic graph SynDG.
	
	%reflect the importance of nodes in the graph
	%We exploit topological centralities of nodes and syntactic features to detect causal syntactic features,
	%And then we select  substitute of casual structure to generate label-changed counterfactuals, which is a syntax-based counterfactual generation method, named \textbf{SynCo}.
	%The centrality of the topological structure ensures the grammatical invariance of substitution. Similar attributes but different labels ensures the quality of counterfactual generation.
	
\begin{figure}[htb]
		\vspace{-1mm}
		\center{\includegraphics[width=0.5\textwidth]{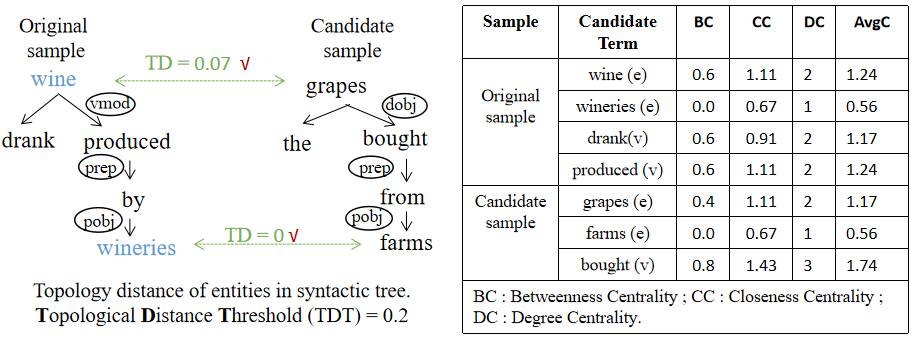}}
		\caption{An illustration of SynCo. The left is SynDG of two samples and the topological distance between entities. The right is the result for centrality calculation. TDT = 0.2.}
		\label{fig:topology}
		\vspace{-4mm}
	\end{figure}

	We take two samples for illustration: an original sample $X_{\text{ori}}$ ``They drank wine produced by wineries.'' with ``Product-Producer'' relation, and a candidate sample $X_{\text{can}}$ ``They bought the grapes from farms'' with ``Entity-Origin''. Their SynDGs and the intervention procedure are shown in Figure~\ref{fig:topology}. %(constructed using a syntactic parsing tool~\citep{ManningSBFBM14})
	The main procedure for SynCo is summarized below.
	
	1. We first calculate the average score for  three centrality metrics of the entity and denote it as avgC(entity), e.g., avgC(wine$_o$) and avgC(grapes$_c$) is 1.24 and 1.17.
	%The average of the three topological centralities for entity `wine' (e$_o$) and `wineries' (e$_o$) in the original sample is 1.24 and 0.56, respectively. And for `grapes' (e$_c$) and `farms' (e$_c$), and root `bought' (r$_c$), TC is 1.17, 0.56, and 1.74, respectively.
	
	2. We then calculate the topological distance (TD) of two entities by minusing their avgC values. If TD is smaller than a predefined topological distance threshold (TDT), they form a candidate entity pair for substitute structures. For example, given TD(avgC(wine$_o$), avgC(grapes$_c$)) =  0.07 $<$ TDT and TD(avgC(wine$_o$), avgC(farms$_c$)) = 0.68 $>$ TDT, the entity `grapes' in $X_{\text{can}}$ and the entity `wine' in $X_{\text{ori}}$  form a candidate entity pair.
	
	3. We now generate the candidate syntactic counterfactual. We calculate the  cosine similarity of the  syntactic features (POS and dependency relation embedding) for  entities in the candidate entity pair. We denote it as FS.
	If it is greater than a predefined feature similarity threshold (FST), we substitute the first-order neighbors around the entity in $X_{\text{ori}}$ with those of the candidate entity in $X_{\text{can}}$. These neighbors should have the same type of POS tags to ensure the correctness of the syntax. Moreover, if there are several words  around entities and they all have the same type, we will replace the one with the smallest TD.
	% And if two verbs around e$_o$ are to be replaced, we will replace with the smallest TD between verbs in e$_c$ and e$_o$. 	%As is shown in Figure~\ref{fig:attribute},
	For example, given FS(wine$_o$, grapes$_c$) = 0.877 $>$ FST,  we can substitute the verb-type neighbors `drank$_o$' and `produced$_o$' of `wine' with the same type neighbor  `bought$_c$' of `grapes'. Moreover, since TD(avgC(drank$_o$), avgC(bought$_c$)) = 0.57 and TD(avgC(produced$_o$), avgC(bought$_c$)) = 0.50, we replace `produced$_o$' with `bought$_c$' and retain `drank$_o$'  unchanged.
	
	%for a word $t_i$,  we use a one-hot vector $b_{pos} \in R^{N_{\text{pos}}}$ and a multi-hot vector $b_{dep} \in R^{N_{\text{dep}}}$ to represent its POS tag and dependency relation(s), where $N_{\text{pos}}$ and $N_{\text{dep}}$ are the number of tag/relation types.
	%If TD is less than the topological distance threshold, the special term in candidate sample is considered as syntactically important term and its first-order neighbor is put into candidite casual strcture pool for $X_{\text{ori}}$. We select `grapes' (e$_c$) rather than `bought' (r$_c$) as syntactically important terms in $X_{\text{ca}}$ corresponding to `wine' in $X_{\text{ori}}$.
	
	%(1) For the entity words `wine' (e$_o$), `wineries' (e$_o$), root word `bought' (R$_o$) and verbs `produced' (v$_o$) in the original sample, the average of the three topological centralities is 1.24, 0.56, 1.17, 1.24. Meanwhile, for the candidate sample, the average of the three topological centralities of the entity words `grapes' (E$_c$), `farms' (E$_c$), and root word bought (R$_c$) is 1.17, 0.56 and 1.74. (2) We calculate TD (wine(E$_o$), grapes(E$_c$)) = 0.07 $<$ TDT,
	%TD (wineries (E$_o$), farms (E$_c$)) = 0 $<$ TDT, TD (bought(R$_o$), bought(R$_c$)) = 0.54 $>$ TDT,
	%TD (bought(R$_c$), produced(V$_o$)) = 1.5 $>$ TDT.
	%(3) we select `grapes' (E$_c$) rather than `bought' (R$_c$) as syntactically important terms in $X_{\text{ca}}$ corresponding to `wine' in $X_{\text{ori}}$. (4) And the first-order neighbors of candidate casual term `grapes' is put into the candidate casual structure pool for original sample $X_{\text{ori}}$.

	4. After substituting the contexts of the candidate entity pair  and adding the label of $X_{\text{can}}$, a candidate counterfactual is produced. In order to ensure that the sample we generate is a real counterfactual, we put it into the trained backbone model and re-predict its label. If the label is indeed changed, we treat it as the counterfactual of $X_{\text{ori}}$.  For example, ``They drank wine bought from wineries.'' with ``Entity-Origin'' relation is a counterfactual sample after SynCo.

	\subsection{Generating Semantic  Counterfactuals}
	%The semantic knowledge of the text is an important factor for relation extraction.
	This section presents our semantically intervened counterfactual generator \textbf{SemCo}. We prefer to replace the contexts between entities in $X_{\text{ori}}$ with words of the most similar semantics and get a different label.
	To this end, we exploit the shortest path in SemDG for identifying semantically similar contexts between entities.
	%We exploit semantic dependency graph of text for identifying semantically important topology.
	
\begin{figure}[htb]
		\vspace{-3mm}
		\center{\includegraphics[width=0.35\textwidth]{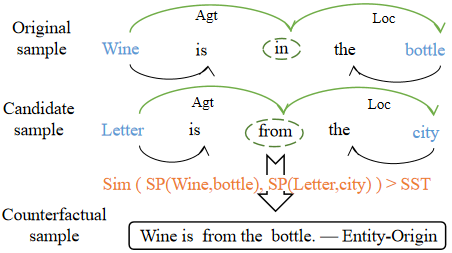}}
		\vspace{-1mm}
		\caption{An illustration of SemCo.  SST = 0.6. SP(e1,e2): the shortest semantic path between two entities.
		}
		\label{fig:semantics}
		\vspace{-1mm}
	\end{figure}

	As shown in Figure~\ref{fig:semantics}, we first obtain the SemDG of an original sample ``Wine is in the bottle'' with ``Content-Container'' label and that of a candidate sample ``Letter is from the city'' with ``Entity-Origin'' label. We then extract the SDP between entities in $X_{\text{ori}}$ and $ X_{\text{can}}$. We propose to calculate the cosine similarity of averaged word embeddings between two paths. If the similarity score is larger than the semantic similarity threshold (SST), we replace the semantic path in  $X_{\text{ori}}$ with that in $X_{\text{can}}$.
	Similar to SynCo, we put the generated semantic counterfactual into the trained backbone model and re-predict its label. We consider the new sample as a counterfactual of $X_{\text{ori}}$ if the label is changed.
	%And then we extract the shortest semantic path between entities and calculate the semantic similarity of two sample's paths.
	%If the two semantic path is similar, we will replace the semantic path in the original sample with another in candidate sample. If the label of new sample is changed, we consider it as counterfactual sample and put it into the counterfactual dataset.
	
	Finally, the generated semantic counterfactuals, as well as the syntactic counterfactuals, are used to augment the original  data to train a robust classifier.
	%\subsection{Training a Robust Classifier}
	%We augment the original training data with the generated semantic counterfactuals to train a robust classifier. We perform experiments below to investigate how do causal syntactic structure and casual semantic path affect the quality of automatically generated counterfactual samples.

	\section{Datasets and Evaluation Protocol}
	\subsection{Datasets}
	We briefly introduce four datasets in our experiments.
	%: three for in-domain test and one for out-of-domain test
	\paragraph{In-Domain Data}
	We  adopt three in-domain datasets on two RC tasks including cross-sentence $n$-ary RC and sentence-level RC. For cross-sentence $n$-ary RC, we use the PubMed dataset which contains 4 relation labels and 1 special \textit{none} label.
	%Most instances contain multiple sentences and each instance is assigned with one of the following five labels.
	%:  ``resistance or non-response'', ``sensitivity'', ``response'', ``resistance'' , and ``none''.
	%Following  \cite{PengPQTY17,SongZWG18}, we consider two sub-tasks for evaluation, i.e., multi-class $n$-ary  RE and  binary-class $n$-ary RE for which we binarize multi-class labels by grouping all relation classes as ``Yes'' and treat ``None'' as ``No''.
	For sentence-level RC task, we follow the experiment settings in \citep{GuoZL19} to evaluate our model on two datasets. (1) SemEval 2010 Task 8 dataset
	%\footnote{\href{ https://github.com/Cartus/AGGCN/tree/master/semeval/dataset}{ https://github.com/Cartus/AGGCN/tree/master/semeval/dataset}.}~\cite{HendrickxKKNSPP10}
	%has 8,000 instances for training and 2,171 instances for testing.
	contains 9 directed relations and 1 \textit{Other} class. (2) TACRED dataset
	%\footnote{\href{https://nlp.stanford.edu/projects/tacred/}{https://nlp.stanford.edu/projects/tacred/}}~\cite{ZhangZCAM17}
	%has over 106k instances and
	contains 41 relation types and 1  \textit{no\_relation} class.
	
	\begin{comment}
	Details of train/validation/test splits for PubMed, TACRED, and Semeval are shown in Table~\ref{DB}.
	\begin{table}[!h]	
	\small	
	\setlength{\abovecaptionskip}{2pt}
	\setlength{\belowcaptionskip}{0pt}
	\footnotesize
	\renewcommand\arraystretch{1.1}
	\centering
	\caption{Train/validation/test splits for PubMed, TACRED and SemEval. ``C'' and ``I'' denotes the number of categories and instances, respectively.}
	\vspace{-3mm}
	\begin{tabular}{lcccccc} \\
	\toprule[1pt]
	\multirow{2}[0]{*}{\textbf{Dataset}} & \multicolumn{2}{c}{\textbf{PubMed}} & \multicolumn{2}{c}{\textbf{TACRED}} & \multicolumn{2}{c}{\textbf{Semeval}}\\
	\cmidrule(r){2-3} \cmidrule(r){4-5}	\cmidrule(r){6-7}
	& C & I & C & I &	C & I\\
	\hline
	Train&	5&	5313 &42 &68124 &10 &8000\\
	\hline
	Validation& 5&200 &42 & 22631&- &-\\
	\hline
	Test&	5&	1474 &42 & 15509&10 &2717\\
	\midrule[1pt]		
	\end{tabular}%
	\vspace{-2mm}
	\label{DB}%
	\end{table}%
	
	\end{comment}
	
	\paragraph{Out-of-Domain Data}
	We employ one  dataset to verify  the model robustness with CAD: the ACE 2005 dataset with 6 relation types and 1 special \textit{none} class.
	%	~\cite{YuGD15}
	It has 6 different domains including broadcast conversation (bc), broadcast news (bn), conversational telephone conversation (cts), newswire (nw), usenet (un), and webblogs (wl).

	\subsection{Evaluation Protocol}
	We evaluate our model with a dedicated protocol. We first adopt three typical RC methods as the backbone, which will be used as the encoder for the input and for training the base classifier for the prediction of the candidate counterfactual. Moreover, the backbone is used for training the final classifier on the original data and CAD.
	The backbone RC methods include PA-LSTM~\citep{ZhangZCAM17}, AGGCN~\citep{GuoZL19}, and R-BERT~\citep{WuH19a}. They are chosen as the representative of sequence based models, graph-based models, and the PLMs methods, respectively.  R-BERT is the best among the state-of-the-art BERT$_\text{base}$ methods. Note the backbone methods will be combined with all counterfactual generation methods including ours and the baselines.

	We further compare our model with automatical counterfactual generation baselines. These methods are either general text generator or designed for other tasks. For the general  methods~\citep{MadaanPPS21,FernP21,RobeerBF21}, we directly generate counterfactuals on RC datasets using their generator, and we avoid masking entities for a fair comparison with our model. Among the specific methods, COSY~\citep{YuZNS020} is for the cross-lingual understanding task, and we follow it to build a counterfactual dependency graph by maintaining the graph structure, and replacing each type of relation with a randomized one and randomly selecting a POS tag. RM/REP-CT~\citep{Yang0CZSD20} is proposed for sentiment classification tasks. We employ its self-supervised context decomposition and select the human-like counterfactuals using MoverScore.

	For models with static word embedding, we initialize word vectors with 300-dimension GloVe embeddings provided by~\cite{PenningtonSM14}.
	For models with contextualized word embedding, we adopt BERT$_\text{base}$ as the PLM.
	The embeddings for POS and dependency relation are initialized randomly and their dimension is set to 30. We adopt the Stanford CoreNLP ~\citep{ManningSBFBM14} as the syntactic parser and use semantic dependency parsing ~\citep{DozatM18} to generate semantic dependency graph.
	We use SGD as the optimizer with a 0.9 decay rate. The L2-regularization coefficient $\lambda$ is 0.002~\footnote{Please refer to Appendix for detailed information for datasets and baselines, as well as the hyper-parameter setting for three thresholds.}.
	
	%Since there is currently no counterfactual generation method used in RE tasks, we have implemented the counterfactual generation method in other areas and employed it on RE tasks. For COSY~\citep{YuZNS020} in cross-lingual understanding tasks, we follow it to build a counterfactual dependency graph by maintaining graph structure and nodes, and replacing each types of relation with a randomized type and randomly selecting a POS tag. We employ self-supervised contextual decomposition and select the human-like counterfactuals using MoverScore like RM/REP-CT~\citep{Yang0CZSD20}. For counterfactual text generator methods GYC~\citep{MadaanPPS21}, CLOSS~\citep{FernP21}, and CounterfactualGAN~\citep{RobeerBF21}, we use these methods on the RE dataset to generate counterfactuals. During the experiment, we found that these methods tend to change the entity words in the original sample.

		\begin{table}[h]
		\vspace{-4.5mm}
		\small
		\centering	
		\vspace{-3mm}
		\renewcommand\arraystretch{1}
		\setlength{\tabcolsep}{0.8mm}{
			\begin{tabular}{lllllll} \\
				\toprule[1pt]
				\multirow{3}[0]{*}{\textbf{Model}} & \multicolumn{4}{c}{\textbf{Binary-class}} & \multicolumn{2}{c}{\textbf{Multi-class}}  \\
				\cmidrule(r){2-5} \cmidrule(r){6-7}
				&\multicolumn{2}{c}{\textbf{T}} & \multicolumn{2}{c}{\textbf{B}} & \textbf{T} & \textbf{B}\\	
				& Single  & Cross  & Single  & Cross   & Cross  & Cross\\
				\hline		
				PA-LSTM$_\text{backbone}$   & 84.9&	 85.8 & 	85.6&	85.0 &	78.1& 	77.0 \\
				COSY &\underline{85.2}&	\underline{86.1}&	\underline{85.8}&	\underline{85.4}&\underline{78.4}&	\underline{77.3}\\
				RM/REP-CT &85.0&	84.9$\downarrow$&	84.9$\downarrow$&	85.0&	77.7$\downarrow$&	76.9$\downarrow$	\\
				GYC &85.1	&85.9&	85.4$\downarrow$&	85.3&	78.2&	77.2	\\
				CLOSS&84.9$\downarrow$&	84.6$\downarrow$&	85.7&	85.2&	77.4$\downarrow$&	76.9$\downarrow$\\
				CounterfactualGAN &84.6$\downarrow$&	85.7$\downarrow$&	85.1$\downarrow$&	85.2&	78.0$\downarrow$&	77.1\\
				%SCIE &-&-&	-&	-&	-&-\\
				%\textbf{SynCo} &86.6&	86.4&	86.8&	86.2&	79.6&	79.4\\
				\textbf{CoCo} &\textbf{87.0}$^\ast$$^\ddagger$	&\textbf{86.9}$\hat{~}$	&\textbf{87.5}$\hat{~}$$^\dagger$	&\textbf{86.8}$^\ast$$^\ddagger$	&\textbf{80.5}$^\ast$$^\ddagger$	&\textbf{80.0}$^\ast$$^\ddagger$
				\\
				\hline	
				AGGCN$_\text{backbone}$   &87.1	&87.0 	&85.2	&85.6 &79.7 &77.4\\
				COSY&\underline{87.6} & \underline{87.8}&\underline{86.3} &\underline{86.2} &\underline{80.7} &\underline{78.4}\\
				RM/REP-CT &87.1 &87.0	 &85.3	 &	85.5$\downarrow$ &	79.7 &77.0$\downarrow$ \\
				GYC &87.2&	87.1&	85.4&	85.6&	79.8&	77.6	\\
				CLOSS&76.4$\downarrow$	&86.3$\downarrow$	&85.0$\downarrow$	&84.3$\downarrow$	&79.2$\downarrow$	&76.4$\downarrow$\\			
				CounterfactualGAN &87.1	&87.1	&85.1&	85.4&	80.1&	77.2\\
				%SCIE &-&-&	-&	-&	-&-\\
				\textbf{CoCo} &\textbf{89.0}$^\ast$$^\ddagger$	&\textbf{89.1}$^\ast$$^\ddagger$&\textbf{88.0}$^\ast$$^\ddagger$&	\textbf{87.7}$^\ast$$^\dagger$&	\textbf{84.1}$^\ast$$^\ddagger$&	\textbf{81.1}$^\ast$$^\ddagger$ \\		
				\hline		
				R-BERT$_\text{backbone}$ & 88.6&	88.7&	88.1&	\underline{87.9}&	85.1&	84.2 \\
				COSY&88.6	&88.8&	\underline{88.3}&	\underline{87.9}&	\underline{85.3}&\underline{84.5}\\
				RM/REP-CT &88.5&88.6$\downarrow$&87.9$\downarrow$&87.6$\downarrow$&	85.0$\downarrow$&84.2\\
				GYC &\underline{88.7}	&\underline{88.9}&	88.0$\downarrow$&	87.8$\downarrow$&	85.2&	84.3	\\
				CLOSS&87.2$\downarrow$	&87.5$\downarrow$	&87.1$\downarrow$&	86.7$\downarrow$&	83.8$\downarrow$&	83.4$\downarrow$\\
				CounterfactualGAN &88.1$\downarrow$	&88.2$\downarrow$	&87.4$\downarrow$	&87.3$\downarrow$	&84.8$\downarrow$&	84.1$\downarrow$\\
				%SCIE &-&-&	-&	-&	-&-\\
				%\textbf{SynCo} &86.6&	86.4&	86.8&	86.2&	79.6&	79.4\\
				%\textbf{SemCo} &
				\textbf{CoCo} &\textbf{89.1}$\hat{~}$	&\textbf{89.3}$\hat{~}$	&\textbf{88.7}$\hat{~}$	&\textbf{88.4}$\hat{~}$	&\textbf{86.2}$\hat{~}$$^\dagger$	&\textbf{85.8}$\hat{~}$$^\dagger$ \\	
				\midrule[1pt]		
		\end{tabular}}
		\vspace{-2mm}
	\caption{Main results in terms of accuracy on PubMed. ``T'' and ``B'' denote the ternary and binary entity interactions, and ``Single'' and ``Cross'' mean the accuracy calculated within single sentences or on all sentences. The best results  are in bold, and the second best ones are underlined. $\hat{~}$ and $\ast$ mark denote statistically significant improvements over the backbone results with $p <$ .05 and $p <$ .01, and $\dagger$  and $\ddagger$  mark denote statistically significant improvements over the corresponding second best results with $p <$ .05 and $p <$ .01, respectively.}
		\label{cross}%
		\vspace{-1mm}
	\end{table}%

\begin{table}[h]	
	\vspace{-4.5mm}
	\small	
	\setlength{\abovecaptionskip}{2pt}
	\setlength{\belowcaptionskip}{0pt}
	\footnotesize
	\renewcommand\arraystretch{1}
	\centering
	
	\vspace{-3mm}
	\setlength{\tabcolsep}{1.2mm}{
		\begin{tabular}{lccll} \\
			\toprule[1pt]
			\multirow{2}[0]{*}{\textbf{Model}} & \multicolumn{3}{c}{\textbf{TACRED}}& \textbf{SemEval}\\
			\cmidrule(r){2-4} \cmidrule(r){5-5}
			& \textbf{P}& \textbf{R}& \textbf{Micro-F1} & \textbf{Macro-F1}\\
			\hline		
			PA-LSTM$_\text{backbone}$  & 65.7  &	64.5&	65.1 & 82.7\\
			COSY &65.8&	64.6&	\underline{65.2}&\underline{83.1}\\
			RM/REP-CT &66.9&63.3$\downarrow$&	65.0$\downarrow$&80.1\\
			GYC & 65.2$\downarrow$&	64.1$\downarrow$&	64.6$\downarrow$& 82.9	\\
			CLOSS&64.2$\downarrow$	&63.9$\downarrow$&	64.0$\downarrow$&81.3$\downarrow$	\\		
			CounterfactualGAN &64.9$\downarrow$	&63.7$\downarrow$&	64.3$\downarrow$&82.4$\downarrow$\\
			%SCIE &-&-&	-&	\\
			\textbf{CoCo} &\textbf{66.3}&	\textbf{66.1}&	\textbf{66.2}$^\ast$$^\dagger$&\textbf{84.2}$^\ast$$^\ddagger$	\\
			\hline	
			AGGCN$_\text{backbone}$  &71.9 &	64.0 &	67.7& 85.7\\
			COSY&71.8$\downarrow$&	64.2&	\underline{67.8}&\underline{85.9} \\
			RM/REP-CT &71.0$\downarrow$	&63.9$\downarrow$&	67.6$\downarrow$&84.8$\downarrow$\\
			GYC & 71.3$\downarrow$&	63.9$\downarrow$&	67.4$\downarrow$& 85.6$\downarrow$	\\
			CLOSS&70.2$\downarrow$&	63.3$\downarrow$&	66.6$\downarrow$&84.8$\downarrow$	\\			
			CounterfactualGAN &71.0$\downarrow$&	63.3$\downarrow$&	66.9$\downarrow$&85.4$\downarrow$\\
			%SCIE &-&-&	-&	\\
			\textbf{CoCo} &\textbf{72.4}&	\textbf{64.8}&	\textbf{68.4}$\hat{~}$$^\dagger$ &\textbf{86.6}$\hat{~}$$^\dagger$\\		
			\hline		
			R-BERT$_\text{backbone}$ &69.7&	70.1&	69.9&\underline{88.6}\\
			COSY &69.6$\downarrow$&	70.1&	69.8$\downarrow$& 88.5$\downarrow$\\
			RM/REP-CT &69.2$\downarrow$	&69.8$\downarrow$&	69.5$\downarrow$&87.2$\downarrow$\\
			GYC &69.5$\downarrow$ &70.5	& \underline{70.0}& \underline{88.6}	\\
			CLOSS&68.0$\downarrow$	&67.9$\downarrow$&	67.9$\downarrow$&87.5$\downarrow$	\\		
			CounterfactualGAN &68.9$\downarrow$	&69.3$\downarrow$&	69.1$\downarrow$&87.5$\downarrow$\\
			%SCIE &-&-&	-&	\\
			\textbf{CoCo} &\textbf{70.2}&	\textbf{70.5}&	\textbf{70.4}&\textbf{89.0} \\	
			\midrule[1pt]		
		\end{tabular}%
	}
	\vspace{2mm}
\caption{Main results on TACRED and  SemEval. Micro-avg. precision (P), recall (R), and F1 on TACRED. Macro-avg. F1 on SemEval. The marks are as same as those in Table 1.}
	\label{sentence}%
	\vspace{-2mm}
\end{table}%

	\section{Results and Discussions}
	We first compare our method with the backbone and the state-of-art automatical  counterfactual generation methods on three in-domain datasets.
	We also provide an ablation study to focus on the contribution of single component in our model.
	Notably, we  evaluate our model on the out-of-domain data for the generalization test. Furthermore, we present a case study of counterfactuals for a detailed comparison of the  counterfactuals generated by baselines and our model.

	\subsection{Main Comparison Results}
	The comparison results for cross-sentence $n$-ary and sentence-level tasks are shown in Table~\ref{cross} and Table~\ref{sentence}, respectively. From these results, we make the following observations.
	
	(1) Our CoCo model significantly improves the performance of all the backbones across three datasets. Specifically, CoCo outperforms two backbone approaches PA-LSTM and AGGCN by around 2-3 absolute percentage points in terms of accuracy on PubMed. It also improves F1 scores  by 1-2 absolute percentage points on TACRED and SemEval. More importantly, our model boosts the  performance of R-BERT  on all three datasets especially on PubMed, which is impressive because it is a very strong backbone, and also because most baselines damage the performance of R-BERT.
	
	(2) Our model achieves the state-of-the-art performance in terms of a counterfactual generator for RC. For example, CoCo outperforms the syntax-based model COSY and the best semantics-based model GYC by 1-3 absolute percentage points. COSY randomly replaces syntactic features which cannot flip the label and makes little sense for RC. CounterfactualGAN and GYC get a slight increase on PubMed with PA-LSTM and AGGCN, but their other results  are unsatisfactory.  The results of RM/REP-CT on three datasets fluctuate. On PubMed, some of them keep the same as the original ones and some of them decline, while on SemEval and TACRED they all decline. The poor results of the baseline counterfactual generators can be due to the lack of grammar constraint and the lack of entity-centric viewpoint. We will illustrate this in our case study.
	%Some counterfactuals generated by RM/REP-CT has non-grammatical norms such as lacking of important predicates.
	%These methods generate counterfactuals that are not grammatically correct and uneasy to read.
	
	%we can easily observe that our CoCo approach consistently outperforms other adapter methods across the four datasets.
	
	%Some works generate a counterfactual by making a set of minimal modifications to the semantics of text that flip the label. But, they do not consider the grammatical validity of the generated counterfactuals.
	
	%(2) Moreover, we can easily observe that our COCO approach consistently outperforms most existing counterfactual data generation methods.

	\begin{table*}[h]	
		\small	
		\setlength{\abovecaptionskip}{2pt}
		\setlength{\belowcaptionskip}{0pt}
		\footnotesize
		\renewcommand\arraystretch{1.1}
		\centering
		%The best results are in bold, and the second best ones with GloVe/BERT embeddings are underlined/italic. $\dagger$  and $\ddagger$  mark statistically significant improvements over the second best results with $p <$ .05 and $p <$ .01, respectively.}
		\setlength{\tabcolsep}{1mm}{
			\begin{tabular}{l|l|l|l}
				\toprule[1.5pt]
				\textbf{Input Sentence} & \textbf{RM/REP-CT} & \textbf{CounterfactualGAN} & \textbf{CoCo}\\
				\hline
				\tabincell{l}{The sisters are teenage\\ \textbf{refugees} from a violent \\ \textbf{home}. — Entity-Origin} &
				\tabincell{l}{The sisters are teenage\\ \textbf{refugees} \sout{from} a violent \\ \textbf{home}. — Other} &
				\tabincell{l}{The sisters are teenage\\ \textbf{refugees} \sout{from} \textcolor{red}{sleep} a violent \\ \textbf{home}. — Member-Collection} &
				\tabincell{l}{The sisters are teenage  \textbf{refugees}\\ \sout{from} \textcolor{red}{lived in} a violent \textbf{home}. \\— Member-Collection}\\
				\hline
				\textbf{Input Sentence} & \textbf{CLOSS} & \textbf{GYC} & \textbf{CoCo}\\
				\hline
				\tabincell{l}{the \textbf{wounds} caused by \\the \textbf{scour ging} and the \\ thorns are almost invisible.\\ — Cause-Effect} &
				\tabincell{l}{the \textbf{wounds} \sout{caused} \textcolor{red}{community} \\by the \textbf{scour ging} and the\\  thorns are almost invisible.\\ — Product-Producer} &
				\tabincell{l}{the \textbf{wounds} \sout{caused by} \textcolor{red}{develop}\\ the \textbf{scour ging} and the thorns \\ are \sout{almost} \textcolor{red}{very} invisible.\\ — Product-Producer} &
				\tabincell{l}{the \textbf{wounds} \sout{caused} \textcolor{red}{produced} \\ by \textbf{scour ging} and the thorns\\   are almost invisible.\\ — Product-Producer}\\
				\toprule[1.5pt]						
			\end{tabular}%
		}
			\caption{Case study. Entities are in bold and the words in red represent newly generated words.
			%- indicates no entity masking.
		}
			\label{case}%
		
	\end{table*}%

\begin{table}[h]	
	\vspace{-2mm}
	\small	
	\setlength{\abovecaptionskip}{2pt}
	\setlength{\belowcaptionskip}{0pt}
	\footnotesize
	\renewcommand\arraystretch{1}
	\vspace{-2mm}
	\setlength{\tabcolsep}{0.6mm}{
		\begin{tabular}{lllll} \\
			\toprule[1pt]
			\multirow{2}[0]{*}{\textbf{Model}} & \textbf{PubMed\_B} & \textbf{PubMed\_T}&\textbf{TACRED}& \textbf{SemEval}\\
			
			& \textbf{Acc.}& \textbf{Acc.}& \textbf{Micro-F1} & \textbf{Macro-F1}\\
			\hline		
			PA-LSTM&	77.0&	78.1 &	65.1&	82.7	\\
			+ SynCo	&\underline{79.4}&	\underline{79.6}&	\underline{65.6}&	\underline{83.9}\\
			+ SemCo	&\textbf{79.7}	&\textbf{79.8}&\textbf{65.7}&	\textbf{84.0}\\
			+ Syn-TED&	77.6&	78.4&	65.2&	83.0\\
			+ Sem-BA&	77.4&	78.3&	65.3&	83.2\\	
			
			%+ CoCo&	\textbf{80.0}&	\textbf{80.5}&	\textbf{66.2}&	\textbf{84.2}\\
			\hline	
			AGGCN&	77.4&	79.7 &	67.7&	85.7\\
			+ SynCo &\underline{80.3}&	\underline{82.8}&	\underline{67.9}&\underline{86.3} \\
			+ SemCo &\textbf{80.7}&\textbf{83.1}&\textbf{68.2}&	\textbf{86.5}	\\	
			+ Syn-TED&	 77.6 &	79.9&	67.7& 	85.8\\						
			+ Sem-BA&	 77.8& 	79.8 &	67.8 &	86.1\\							
			%+ CoCo &\textbf{81.1}&	\textbf{84.1}&	\textbf{68.4}&	\textbf{86.6}\\
			
			\hline		
			R-BERT&	84.2&	85.1&	69.9&\underline{88.6}\\
			+ SynCo &\underline{85.1}	&\underline{85.6}&	\underline{70.1}&	\textbf{88.7}\\
			+ SemCo &\textbf{85.6}	&\textbf{85.8}&\textbf{70.3}&	\textbf{88.7}\\
			+ Syn-TED&	84.3& 	85.1 &	70.0 &	88.5\\			
			+ Sem-BA	& 84.6& 85.2&	70.0& 	88.3\\				
			%+ CoCo &\textbf{85.8}	&\textbf{86.2}	&\textbf{70.4}	&\textbf{89.0} \\
			\midrule[1pt]		
		\end{tabular}%
	}
		\centering
	\caption{Ablation results. (\_B) and (\_T)  denote the binary and ternary relation on PubMed.}
	\label{ablation}%
	\vspace{-2mm}
\end{table}%

	\begin{table}[h]	
	\vspace{-2mm}
	\small	
	\setlength{\abovecaptionskip}{2pt}
	\setlength{\belowcaptionskip}{0pt}
	\footnotesize
	\renewcommand\arraystretch{1}
	\centering
	%		\caption{Micro-avg. F1 scores of the models on the ACE 2005 dataset over different target domains bc, cts, and wl.}
	\vspace{-2mm}
	\setlength{\tabcolsep}{1mm}{
		\begin{tabular}{llll} \\
			\toprule[1pt]
			\multirow{2}{*}{\tabincell{l}{\textbf{Different} \\\textbf{Training Data}}}
			& & &\\
			& \textbf{PA-LSTM} & \textbf{AGGCN} & \textbf{R-BERT}\\
			\hline
			\multicolumn{4}{c}{Micro-avg. F1 on \textbf{bc} domain.}\\
			\hline
			Ori. &48.5&	62.5&	68.5\\			
			Ori. \& CAD (GYC) &48.7&	63.1&	68.6\\
			Ori. \& CAD (COSY) &48.6&	62.4 $\downarrow$ &	68.6\\
			Ori. \& CAD (CoCo)&	\textbf{52.6}$^\ast$&	\textbf{64.4}$^\ast$&	\textbf{69.0}$^\ast$\\
			\hline
			\multicolumn{4}{c}{Micro-avg. F1 on \textbf{cts} domain.}\\
			\hline
			Ori. &42.5	&63.1&	69.4\\				
			Ori. \& CAD (GYC) &42.6	&63.8&	69.8\\
			Ori. \& CAD (COSY) &42.3 $\downarrow$&	63.2&	69.7\\
			Ori. \& CAD (CoCo)&	\textbf{46.2}$^\ast$&	\textbf{65.3}$^\ast$&	\textbf{70.4}$\hat{~}$\\
			
			\hline
			\multicolumn{4}{c}{Micro-avg. F1 on \textbf{wl} domain.}\\
			\hline
			Ori. &38.8	&53.4&	59.5\\
			Ori. \& CAD (GYC) &39.1	&53.6&	59.7\\
			Ori. \& CAD (COSY)  &38.1 $\downarrow$&	53.7&	59.2 $\downarrow$\\				
			Ori. \& CAD (CoCo)&	\textbf{41.1}$^\ast$&\textbf{55.2}$^\ast$&\textbf{60.2}$\hat{~}$\\
			\midrule[1pt]		
		\end{tabular}%
	}
		\caption{Results for the generalization test on ACE 2005 dataset.}
	
	\label{robust1}%
	\vspace{-3mm}
\end{table}%

	\subsection{Ablation Study}
	Our model has two unique characteristics. Firstly, it utilizes both syntactic and sematic information. Secondly, it takes advantage of the graph topological property. We hence design two types of ablation study. One is performing one separate component only, i.e., SynCo or SemCo. The other is replacing our graph topology based intervention methods with other syntactic/semantic  ones, i.e., tree edit distance (TED)~\citep{ZhangS89} which measures the syntactic closeness between  the candidate and the original text  and BERT-attack (BA)~\citep{LiMGXQ20} which generates substitutes for the vulnerable words in a semantic-preserving way.
	%TED exploits tree-edit distance to measure the syntactic closeness of candidate and original text.
	%To prove the benefit our our proposed SynCO, we find replace SynCO with TED to casual syntactic structure and generate counterfactual samples.
	%Textual adversaries also aim to change the model prediction (with modifications resembling natural text to ensure the neural networks are highly reliable and robust..
	%BA generates adversarial sample using BERT in a semantic-preserving way to generate substitutes for the vulnerable words.
	%We expliot BERT-attack for RE dataset to generate adversarial samples and add them into our original data to train our backbone.
	We present the ablation results on three datasets in Table~\ref{ablation}. 
	
	%We find both the removal and replacement operations incur the performance decrease in comparison with their counterparts, which clearly demonstrates that our SynCo and SemCo strategy both contribute to our model.
	
	We find that a single SynCo or SemCo is already good enough to enhance the performance of the backbone.
	In addition, we observe that our single SynCo/SemCo outperforms the baselines in Table~\ref{cross} and Table~\ref{sentence}. 	
Moreover, SynCo and SemCo have almost the same effects on the model. For example, the backbone PA-LSTM has an accuracy score 77.0 on PubMed\_B. After SynCo/SemCo, its accuracy rises up to 79.4/79.7, showing a similar 2.4/2.7 absolute increase. %On other datasets, the employment of SynCo or SemCo also has very positive impacts on  backbones' performance.
	These results demonstrate that both SynCo and SemCo contribute to our model, and their combination CoCo is more powerful.

	The replacement of SynCo with TED and SemCo with BA hurts the performance. For example, on SemEval with AGGCN as the backbone, TED results in a 0.5 F1 decrease (SynCo 86.3 \textit{vs}. Syn-TED 85.8). On TACRED with R-BERT  as the backbone, BA brings about a 0.3 (SemCo 70.3 \textit{vs}. Syn-TED 70.0) absolute decrease. The reason might be that TED only considers the causal words from the whole syntactic tree while our SynCo exploits the topological structure and syntactic feature of the words. Meanwhile,  BA is unable to capture casual associations  when generating adversarial samples since it only replaces the words with similar ones generated by BERT.

	\subsection{Robustness in the Generalization Test}
	To evaluate the model robustness, we perform the generalization test on ACE2005 dataset by using the training and test data from different domains.
	Following the settings for fine-grained RC tasks~\citep{YuGD15}, we use the union of the news domains (nw and bn) for training, and hold out half of the bc domain as development data, and finally evaluate on the remainder of bc, cts, and wl domains. We choose two best baselines GYC and COSY for comparison.
	
	%For ACE2005 dataset, following the previous work~\citep{YuGD15} settings, we use union of the news domains (nw and bn) as training data, hold out half of the bc domain as development data, and evaluate on the remainder of bc, cts, and wl. An advantage of the ACE 2005 dataset is it helps to evaluate the cross-domain generalization of the models as the training data and test data in this case comes from different domains.
	%One advantage of the ACE 2005 dataset is that it helps to evaluate the cross-domain generalization of the model, because in this case the training data and test data come from different domains.
	As can be seen in Table~\ref{robust1}, our proposed CoCo model enhances the performance of all backbones on the  combined CAD and the original (Ori.) data, i.e., PA-LSTM (+4.1\%, +3.7\%, +2.3\% ), AGGCN (+1.9\%, +2.2\%, +1.8\%), and R-BERT (+0.5\%, +1.0\%, +0.7\%) on three different target domains. It is worthy of noting that our improvements over three backbones (including R-BERT) are statistically significant on all generalization tests.
	%all of the backbones trained on original (Ori.) dataset and CAD can outperform training on the Ori. dataset varying with the models as follows: PA-LSTM (+4.1\%, +3.7\%, +2.3\% ), AGGCN (+1.9\%, +2.2\%, +1.8\%), and R-BERT (+0.5\%, +1.0\%, +0.7\%) over three different target domains.
	In contrast, two  baselines GYC and COSY only get very small improvements or even decrease the performance of backbones in some cases. All these clearly prove that our method can enhance the robustness of backbone methods with the generated  counterfactuals.

	\subsection{Case Study}	
	
	%To have a close look, we select two input examples from SemEval for a case study and present the results of different models in Table~\ref{case}. For the first sample ``The sisters are teenage refugees from a violent  home'', the counterfactual generated by RM/REP-CT loses the important predicate `from'. The one by CounterfactualGAN has grammar error.	For the second sample ``the wounds  caused by the scour ging and the  thorns are almost invisible'', the counterfactual ``...wounds (noun) community (noun) by (prep)...'' generated by CLOSS  does not conform to the language criterion, and the one generated by GYC is also unreadable.
	
	%Overall, the counterfactuals generated by other methods are very uncontrollable, and it is easy for them to generate sentences with grammar errors. In contrast, the counterfactuals generated by our CoCo model are syntactically correct and fluent in semantics.

To have a close look, we select two samples from SemEval for case study and present results by different models in Table~\ref{case}.
RM/REP-CT deletes the important preposition `from' to change the label and results in an incomplete sample.
CounterfactualGAN employs an adversarial attack method. Its generated `sleep' is ungrammatical and the sample is low-quality as it is inconsistent with the flipped label.
%generates reasonable word `sleep' between `refugees' and `home',  but does not consider whether the `sleep' is grammatical. identify the causal words `caused' but generate not human-like words. They are easily generate related words with given labels, but make the whole sentence unreasonable.
GYC and CLOSS generate the counterfactual  with a specified target label and the substitution tends to be common words in the specified class like `community', which  makes the sentence unreadable.
Moreover,  the word `almost' identified by GYC is out of the scope of two entities and is not a causal feature.

Overall, the results by other methods are very uncontrollable, and it is easy for them to generate sentences with semantic or syntactic errors.
In contrast, our model not only identifies the correct causal features, but also produces human-like counterfactuals, which can force the classifier to better distinguish between causation and confounding.
%the counterfactuals generated by our model are syntactically correct and fluent in semantics. Not only do we find causal features, but we generate more plausible counterfactual sentences, which are more human-like.

	%changes on input text to change the label of output sentence, regardless of whether the output sentence makes sense. Our method CoCo generate counterfactuals text samples with plausibility, label-flip.
	
	%Few examples of given input sentences show that the existing methods generate hard test cases with either single word generation or by adding an existing template to generate test-cases which do not possess the required generation quality in nature. Approaches like GYC and CounterfactualGAN generate changes on input text to change the label of output sentence, regardless of whether the output sentence makes sense.

	\section{Conclusion}
	In this paper, we introduce the problem of automatic counterfactual generation into the RC task. We aim to produce the most human-like,  i.e., grammatically correct and semantically readable,  counterfactuals,   while keeping the entities unchanged. To this end, we design an entity-centric framework which employs semantic and syntactic dependency graphs and exploits two  topological properties in these two graphs to  first identify and then intervene on contextual casual features for entities. Extensive experimental results on  four datasets  prove that our model significantly outperforms the state-of-the-art baselines, and is more effective for alleviating spurious associations and improving the model robustness.
	%\section*{Acknowledgments}
	\clearpage
	\appendix
	\bibliographystyle{named}
	\bibliography{ijcai22}

\begin{thebibliography}{}

\bibitem[\protect\citeauthoryear{Chen \bgroup \em et al.\egroup
  }{2021}]{ChenXY21}
Hao Chen, Rui Xia, and Jianfei Yu.
\newblock Reinforced counterfactual data augmentation for dual sentiment
  classification.
\newblock In {\em EMNLP}, 2021.

\bibitem[\protect\citeauthoryear{Dozat and Manning}{2018}]{DozatM18}
Timothy Dozat and Christopher~D. Manning.
\newblock Simpler but more accurate semantic dependency parsing.
\newblock In {\em ACL}, 2018.

\bibitem[\protect\citeauthoryear{Fern and Pope}{2021}]{FernP21}
Xiaoli~Z. Fern and Quintin Pope.
\newblock Text counterfactuals via latent optimization and shapley-guided
  search.
\newblock In {\em EMNLP}, 2021.

\bibitem[\protect\citeauthoryear{Guo \bgroup \em et al.\egroup
  }{2019}]{GuoZL19}
Zhijiang Guo, Yan Zhang, and Wei Lu.
\newblock Attention guided graph convolutional networks for relation
  extraction.
\newblock In {\em ACL}, 2019.

\bibitem[\protect\citeauthoryear{Jo and Bengio}{2017}]{abs-1711-11561}
Jason Jo and Yoshua Bengio.
\newblock Measuring the tendency of cnns to learn surface statistical
  regularities.
\newblock {\em CoRR}, abs/1711.11561, 2017.

\bibitem[\protect\citeauthoryear{Judea}{2000}]{Judea2000}
Pearl Judea.
\newblock Cambridge University Press, 2000.

\bibitem[\protect\citeauthoryear{Kaushik \bgroup \em et al.\egroup
  }{2020}]{KaushikHL20}
Divyansh Kaushik, Eduard~H. Hovy, and Zachary~Chase Lipton.
\newblock Learning the difference that makes {A} difference with
  counterfactually-augmented data.
\newblock In {\em ICLR}, 2020.

\bibitem[\protect\citeauthoryear{Li \bgroup \em et al.\egroup
  }{2020}]{LiMGXQ20}
Linyang Li, Ruotian Ma, Qipeng Guo, Xiangyang Xue, and Xipeng Qiu.
\newblock {BERT-ATTACK:} adversarial attack against {BERT} using {BERT}.
\newblock In {\em EMNLP}, 2020.

\bibitem[\protect\citeauthoryear{Madaan \bgroup \em et al.\egroup
  }{2021}]{MadaanPPS21}
Nishtha Madaan, Inkit Padhi, Naveen Panwar, and Diptikalyan Saha.
\newblock Generate your counterfactuals: Towards controlled counterfactual
  generation for text.
\newblock In {\em AAAI}, 2021.

\bibitem[\protect\citeauthoryear{Mandya \bgroup \em et al.\egroup
  }{2020}]{MandyaBC20}
Angrosh Mandya, Danushka Bollegala, and Frans Coenen.
\newblock Graph convolution over multiple dependency sub-graphs for relation
  extraction.
\newblock In {\em COLING}, 2020.

\bibitem[\protect\citeauthoryear{Manning \bgroup \em et al.\egroup
  }{2014}]{ManningSBFBM14}
Christopher~D. Manning, Mihai Surdeanu, John Bauer, Jenny~Rose Finkel, Steven
  Bethard, and David McClosky.
\newblock The stanford corenlp natural language processing toolkit.
\newblock In {\em ACL}, 2014.

\bibitem[\protect\citeauthoryear{Noh and Kavuluru}{2021}]{NohK21}
Jiho Noh and Ramakanth Kavuluru.
\newblock Joint learning for biomedical {NER} and entity normalization:
  encoding schemes, counterfactual examples, and zero-shot evaluation.
\newblock In {\em BCB}, pages 55:1--55:10, 2021.

\bibitem[\protect\citeauthoryear{Pennington \bgroup \em et al.\egroup
  }{2014}]{PenningtonSM14}
Jeffrey Pennington, Richard Socher, and Christopher~D. Manning.
\newblock Glove: Global vectors for word representation.
\newblock In {\em EMNLP}, 2014.

\bibitem[\protect\citeauthoryear{Qin \bgroup \em et al.\egroup
  }{2021}]{QinLT00JHS020}
Yujia Qin, Yankai Lin, Ryuichi Takanobu, Zhiyuan Liu, Peng Li, Heng Ji, Minlie
  Huang, Maosong Sun, and Jie Zhou.
\newblock {ERICA:} improving entity and relation understanding for pre-trained
  language models via contrastive learning.
\newblock In {\em ACL}, 2021.

\bibitem[\protect\citeauthoryear{Robeer \bgroup \em et al.\egroup
  }{2021}]{RobeerBF21}
Marcel Robeer, Floris Bex, and Ad~Feelders.
\newblock Generating realistic natural language counterfactuals.
\newblock In {\em EMNLP Findings}, 2021.

\bibitem[\protect\citeauthoryear{Srivastava \bgroup \em et al.\egroup
  }{2020}]{SrivastavaHL20}
Megha Srivastava, Tatsunori~B. Hashimoto, and Percy Liang.
\newblock Robustness to spurious correlations via human annotations.
\newblock In {\em ICML}, 2020.

\bibitem[\protect\citeauthoryear{Tucker \bgroup \em et al.\egroup
  }{2021}]{TuckerQL21}
Mycal Tucker, Peng Qian, and Roger Levy.
\newblock What if this modified that? syntactic interventions with
  counterfactual embeddings.
\newblock In {\em Findings of the ACL}, 2021.

\bibitem[\protect\citeauthoryear{Wang and Culotta}{2021}]{WangC21}
Zhao Wang and Aron Culotta.
\newblock Robustness to spurious correlations in text classification via
  automatically generated counterfactuals.
\newblock In {\em AAAI}, 2021.

\bibitem[\protect\citeauthoryear{Wu and He}{2019}]{WuH19a}
Shanchan Wu and Yifan He.
\newblock Enriching pre-trained language model with entity information for
  relation classification.
\newblock In {\em CIKM}, 2019.

\bibitem[\protect\citeauthoryear{Wu \bgroup \em et al.\egroup }{2021}]{WuRHW20}
Tongshuang Wu, Marco~T{\'{u}}lio Ribeiro, Jeffrey Heer, and Daniel~S. Weld.
\newblock Polyjuice: Generating counterfactuals for explaining, evaluating, and
  improving models.
\newblock In {\em ACL}, 2021.

\bibitem[\protect\citeauthoryear{Xu \bgroup \em et al.\egroup
  }{2015}]{XuMLCPJ15}
Yan Xu, Lili Mou, Ge~Li, Yunchuan Chen, Hao Peng, and Zhi Jin.
\newblock Classifying relations via long short term memory networks along
  shortest dependency paths.
\newblock In {\em EMNLP}, 2015.

\bibitem[\protect\citeauthoryear{Yamada \bgroup \em et al.\egroup
  }{2020}]{YamadaASTM20}
Ikuya Yamada, Akari Asai, Hiroyuki Shindo, Hideaki Takeda, and Yuji Matsumoto.
\newblock {LUKE:} deep contextualized entity representations with entity-aware
  self-attention.
\newblock In {\em EMNLP}, 2020.

\bibitem[\protect\citeauthoryear{Yang \bgroup \em et al.\egroup
  }{2021}]{Yang0CZSD20}
Linyi Yang, Jiazheng Li, Padraig Cunningham, Yue Zhang, Barry Smyth, and Ruihai
  Dong.
\newblock Exploring the efficacy of automatically generated counterfactuals for
  sentiment analysis.
\newblock In {\em ACL}, 2021.

\bibitem[\protect\citeauthoryear{Yu \bgroup \em et al.\egroup }{2015}]{YuGD15}
Mo~Yu, Matthew~R. Gormley, and Mark Dredze.
\newblock Combining word embeddings and feature embeddings for fine-grained
  relation extraction.
\newblock In {\em NAACL}, 2015.

\bibitem[\protect\citeauthoryear{Yu \bgroup \em et al.\egroup
  }{2021}]{YuZNS020}
Sicheng Yu, Hao Zhang, Yulei Niu, Qianru Sun, and Jing Jiang.
\newblock {COSY:} counterfactual syntax for cross-lingual understanding.
\newblock In {\em ACL}, 2021.

\bibitem[\protect\citeauthoryear{Zeng \bgroup \em et al.\egroup
  }{2020}]{ZengLZZ20}
Xiangji Zeng, Yunliang Li, Yuchen Zhai, and Yin Zhang.
\newblock Counterfactual generator: {A} weakly-supervised method for named
  entity recognition.
\newblock In {\em EMNLP}, 2020.

\bibitem[\protect\citeauthoryear{Zhang and Shasha}{1989}]{ZhangS89}
Kaizhong Zhang and Dennis~E. Shasha.
\newblock Simple fast algorithms for the editing distance between trees and
  related problems.
\newblock {\em {SIAM} J. Comput.}, 1989.

\bibitem[\protect\citeauthoryear{Zhang \bgroup \em et al.\egroup
  }{2017}]{ZhangZCAM17}
Yuhao Zhang, Victor Zhong, Danqi Chen, Gabor Angeli, and Christopher~D.
  Manning.
\newblock Position-aware attention and supervised data improve slot filling.
\newblock In {\em EMNLP}, 2017.

\bibitem[\protect\citeauthoryear{Zhou \bgroup \em et al.\egroup
  }{2016a}]{Zhou_cvpr16}
Bolei Zhou, Aditya Khosla, Agata Lapedriza, Aude Oliva, and Antonio Torralba.
\newblock Learning deep features for discriminative localization.
\newblock In {\em CVPR}, 2016.

\bibitem[\protect\citeauthoryear{Zhou \bgroup \em et al.\egroup
  }{2016b}]{ZhouSTQLHX16}
Peng Zhou, Wei Shi, Jun Tian, Zhenyu Qi, Bingchen Li, Hongwei Hao, and Bo~Xu.
\newblock Attention-based bidirectional long short-term memory networks for
  relation classification.
\newblock In {\em ACL}, 2016.

\bibitem[\protect\citeauthoryear{Zhu \bgroup \em et al.\egroup
  }{2020}]{ZhuZLW20}
Qingfu Zhu, Wei{-}Nan Zhang, Ting Liu, and William~Yang Wang.
\newblock Counterfactual off-policy training for neural dialogue generation.
\newblock In {\em EMNLP}, 2020.

\end{thebibliography}
	
\end{document}